\pdfoutput=1

\documentclass[11pt]{article}

\usepackage[final]{acl}

\usepackage{times}
\usepackage{latexsym}
\usepackage{adjustbox}
\usepackage{enumitem}
\usepackage{booktabs}
\usepackage[T1]{fontenc}
\usepackage{natbib}
\usepackage[utf8]{inputenc}

\usepackage{microtype}
\usepackage{multicol}
\usepackage{multirow}
\usepackage{multirow}
\usepackage{makecell}
\usepackage{array}
\usepackage{booktabs}
\usepackage{algorithm}
\usepackage{algpseudocode}
\usepackage{amsmath}
\usepackage{colortbl}
\usepackage{hhline}
\usepackage{inconsolata}
\usepackage{marginnote}
\usepackage{graphicx}
\usepackage{amssymb}
\usepackage{pifont}
\usepackage[flushmargin]{footmisc} 
\title{PhysReason: A Comprehensive Benchmark towards \\ Physics-Based Reasoning}

\author{Xinyu Zhang$^{1,4}$,\; Yuxuan Dong$^{1,4}$,\; Yanrui Wu$^{1,4}$,\; Jiaxing Huang$^{1,4}$,\;  Chengyou Jia $^{1,4}$,\; \\ \textbf{Basura Fernando} $^{2}$,\; \textbf{Mike Zheng Shou} $^{3}$,\;  
\textbf{Lingling Zhang}$^{1,4}$ \thanks{Corresponding author}, \; 
\textbf{Jun Liu}$^{1,5}$
\\
{$^{1}$School of Computer Science and Technology, Xi’an Jiaotong University}\; \\
{$^{2}$Institute of High-Performance Computing, Agency for Science, Technology and Research, Singapore} \; \\
{$^{3}$Show Lab, National University of Singapore} \; \\
{$^{4}$Ministry of Education Key Laboratory of Intelligent Networks and Network Security, China} \; \\
{$^{5}$Shaanxi Province Key Laboratory of Big Data Knowledge Engineering, China} \; \\
\texttt{{zhang1393869716}@stu.xjtu.edu.cn, \{zhanglling,liukeen\}@xjtu.edu.cn}
}

\begin{document}
\maketitle
\begin{abstract}
Large language models demonstrate remarkable capabilities across various domains, especially mathematics and logic reasoning.
However, current evaluations overlook physics-based reasoning, a complex task requiring physics theorems and constraints.
We present PhysReason, a 1,200-problem benchmark comprising knowledge-based (25\%) and reasoning-based (75\%) problems, where the latter are divided into three difficulty levels (easy, medium, hard).
Notably, problems require an average of 8.1 solution steps, with hard problems requiring 15.6, reflecting the complexity of physics-based reasoning.
We propose the Physics Solution Auto Scoring Framework, incorporating efficient answer-level and comprehensive step-level evaluations.
Top-performing models like Deepseek-R1, Gemini-2.0-Flash-Thinking, and o3-mini-high achieve less than 60\% on answer-level evaluation, with performance dropping from knowledge questions (75.11\%) to hard problems (31.95\%).
Through step-level evaluation, we identify four key bottlenecks: Physics Theorem Application, Physics Process Understanding, Calculation, and Physics Condition Analysis.
These findings position PhysReason as a novel and comprehensive benchmark for evaluating physics-based reasoning capabilities in large language models.
Our code and data will be published at \url{https://dxzxy12138.github.io/PhysReason/}.
\end{abstract}

\section{Introduction}
Large Language Models (LLMs) have demonstrated remarkable performance across various domains, such as math \cite{lightmanlet, cobbe2021training} and logical reasoning \cite{hendrycksmeasuring, xu2025large, 10584140}. 
However, current evaluations often overlook physics-based reasoning, limiting their applications in scenarios such as robotics \cite{chow2025physbench} and autonomous driving \cite{huang2023applications}. 
This is because physics-based reasoning, integrating multiple theorems and physics constraints, is more closely aligned with practical applications than math and logical reasoning. 
Consequently, developing a comprehensive benchmark for evaluating LLMs' physics-based reasoning capabilities is crucial for discovering current limitations and guiding future improvements.
\par
There are several pioneering physics benchmarks (K-12 level like ScienceQA \cite{lu2022learn}, college-level SciBench \cite{wangscibench}, and expert-level GPQA \cite{rein2024gpqa}) encompassing progressively advanced knowledge domains.
However, they exhibit two critical limitations: oversimplified reasoning processes and neglecting step-level evaluation.
These problems typically involve only 3-4 physics formulas, focusing solely on final answers to measure model performance.
Therefore, a benchmark featuring in-depth reasoning processes and step-level evaluation is urgently needed to measure LLMs' physics-based reasoning capabilities.
\par
\begin{figure*}[t]
\centering
\includegraphics[width=0.96\textwidth]{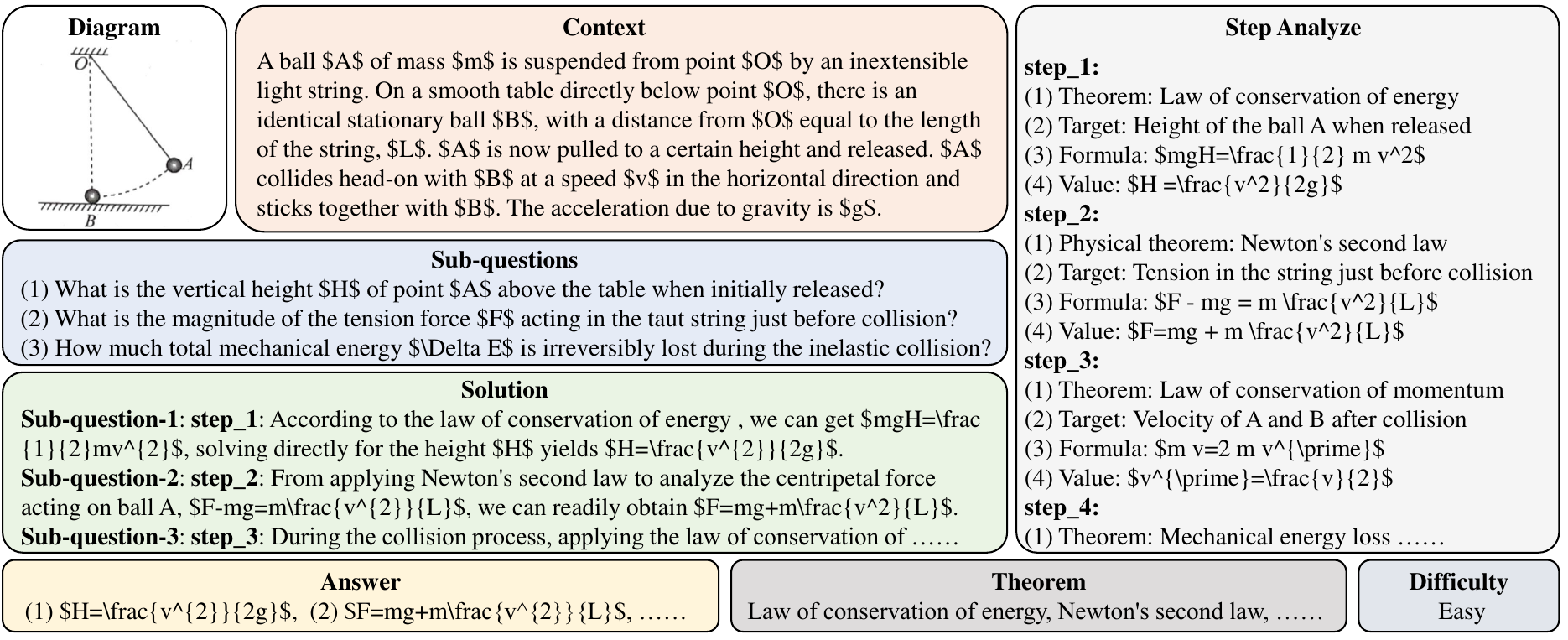}
\vspace{-6pt}
\caption{An illustration of example from our PhysReason benchmark. Due to space constraints, only key components are shown. Please refer to Appendix \ref{Example} for complete annotations.}
\label{fig:0}
\vspace{-15pt}
\end{figure*}
\par
To address these limitations, we present PhysReason, a comprehensive benchmark comprising 1,200 problems designed to evaluate models' physics-based reasoning capabilities. 
As illustrated in Figure \ref{fig:0}, PhysReason features physics problems that require multi-step reasoning and precise application of physics theorems. 
The benchmark introduces several key characteristics:
\begin{enumerate}[leftmargin=*,itemsep=0pt,parsep=0pt,topsep=0pt]
\item \textbf{Stratified difficulty}: There are knowledge-based (25\%) and reasoning-based (75\%) problems, with reasoning problems categorized into easy, medium, and hard (25\% each).
\item \textbf{Complex reasoning}: Solutions average 8.1 steps per problem, with hard problems reaching 15.6 steps, exceeding current physics benchmarks which typically only contain 3-4 steps.
\item \textbf{Multi-modal design}: 81\% of problems include diagrams, evaluating models' capabilities in comprehending visual and textual information.
\end{enumerate}
\par
To evaluate performance on PhysReason comprehensively, we propose the Physics Solution Auto Scoring Framework (PSAS) based on current LLMs' capabilities in information extraction and formula comparison. 
This framework encompasses two answer-level and step-level evaluation methods, PSAS-A and PSAS-S.
PSAS-A enables efficient evaluation through answer comparison, while PSAS-S facilitates comprehensive analysis through step-by-step reasoning verification. 
Experimental results demonstrate that PSAS significantly outperforms direct LLM-based evaluation approaches, achieving an evaluation accuracy exceeding 98\%.
\par
We evaluate seven non-O-like models and eight O-like models on the PhysReason benchmark. 
Results show that while Deepseek-R1 \cite{guo2025deepseek}, Gemini-2.0-Flash-Thinking-0121 \cite{google_gemini_thinking}, and o3-mini-high \cite{o3_mini} demonstrate superior performance, their average scores remain below 60\%. 
Moreover, models excel in basic physics concepts but consistently show performance degradation as problem difficulty and required solution steps increase (from 75.11\% to 31.95\%). 
This degradation stems from the models' inability to maintain accuracy across consecutive solution steps, so maintaining the reliability of the reasoning process is crucial. 
Through step-level evaluation, we identify four critical bottlenecks limiting model performance: Physics Theorem Application, Physics Process Understanding, Calculation Process, and Physics Condition Analysis.
\section{Related Work}
\subsection{Large Language Model Evaluation}
LLMs have demonstrated remarkable performance across various domains, including math reasoning \cite{jiang2024forward, li2024snapkv, imani2023mathprompter}, logical reasoning \cite{sun2024determlr, xu-etal-2024-symbol, 10836885}, and text generation \cite{zhao2024docmath, liang2024controllable}. 
However, these models exhibit notable limitations when confronted with physics-based interactions, significantly constraining their deployment in autonomous driving and robotics applications \cite{gao2024physically}. 
Unlike mathematical and logical reasoning, physics-based reasoning requires the sophisticated integration of multiple principles alongside real-world physical constraints \cite{kline1981mathematics}. 
Consequently, developing robust physics reasoning capabilities represents a crucial prerequisite for expanding LLMs' potential in practical scenarios \cite{lai2024vision}. 
Current evaluation methodologies predominantly focus on math or logical reasoning domains, revealing a critical gap in systematically assessing LLMs' physics reasoning proficiency.

\subsection{Physics Benchmarks}
Existing physics benchmarks span three knowledge complexity levels: K-12 (ScienceQA \cite{lu2022learn}, E-EVAL \cite{hou-etal-2024-e}), college-level (MMLU \cite{hendrycksmeasuring}, AGIEval \cite{zhong2024agieval}, JEEBench \cite{arora2023have}, TheoremQA \cite{chen2023theoremqa}, EMMA \cite{hao2025can}, SciEval \cite{sun2024scieval}, C-Eval-STEM \cite{huang2024c}, SciBench \cite{wangscibench}), and expert-level (OlympiadBench\cite{he-etal-2024-olympiadbench}, GPQA \cite{rein2024gpqa}).
While these benchmarks showcase LLMs' knowledge breadth, they simplify reasoning to 3-4 steps and emphasize only final answers.
PhysReason addresses these gaps through complex reasoning process and step-level evaluation.
\begin{table*}
\centering
\caption{Comparative analysis of our PhysReason with other physics-based reasoning benchmarks.
For \textbf{Knowledge}, COMP: Competition, COL: College, CEE: College Entrance Examination, K1-K12: Elementary and High School, PH.D: Doctor of Philosophy;
For \textbf{question type}, OE: Open-ended, MC: Multiple-choice, Avg. T: Average Tokens;
For \textbf{solution type}, Avg. S: Average Steps.}
\vspace{-5pt}
\begin{adjustbox}{width=\textwidth}
\begin{tabular}{lcccccc@{\hspace{5pt}}ccc}
\hline
\multirow{2}{*}{Benchmark} & \multirow{2}{*}{Multi-modal} & \multirow{2}{*}{Size} & \multirow{2}{*}{Knowledge} & \multicolumn{2}{c}{Question} & \multicolumn{3}{c}{Solution} \\
\cmidrule(r){5-6} \cmidrule(l){7-9}
  &        &           &      & Type & Avg. T & Step-by-step & Avg. T  & Avg. S\\
\hline
JEEBench       & {\color{red}\ding{55}}  & 123  & CEE & OE,MC & 169.7 & -  & - & -\\
MMLU-Pro       &  {\color{red}\ding{55}} & 1299 & COL & MC    & 52.1  & -   & - & -\\
GPQA           & {\color{red}\ding{55}} & 227  & PH.D. & OE    & 111.4 & {\color{red}\ding{55}} & 197.2 & 3.6\\
SciEval        & {\color{red}\ding{55}} & 1657 & - & OE,MC & 154.5 & -  & - & -\\
SciBench       & \color{green}\ding{51} & 295  & COL & OE    & 80.5  & {\color{red}\ding{55}} & 315.9 & 2.8\\
MMMU           & \color{green}\ding{51} & 443  & COL & OE,MC & 53.8  & - & - & -\\
ScienceQA      & \color{green}\ding{51} & 617  & K1-K12 & MC    & 13.3  & {\color{red}\ding{55}} & 63.0 & 2.4\\
OlympiadBench  & \color{green}\ding{51} & 2334 & COMP & OE    & 222.0 & {\color{red}\ding{55}} & 199.8 & 3.7\\
EMMA           & \color{green}\ding{51} & 156  & - & MC    & 109.5 & -  & - & -\\
\hline
Ours-Knowledge  & \color{green}\ding{51} & 300  & CEE+COMP & OE    & 163.7 & \color{green}\ding{51} & 196.5 & 3.3\\
Ours-Easy       & \color{green}\ding{51} & 300  & CEE+COMP & OE    & 171.2 & \color{green}\ding{51} & 241.5 & 5.0 \\
Ours-Medium     & \color{green}\ding{51} & 300  & CEE+COMP & OE    & 229.2 & \color{green}\ding{51} & 391.3 & 8.4\\
Ours-Hard  & \color{green}\ding{51} & 300  & CEE+COMP & OE    & 340.9 & \color{green}\ding{51} & 936.1 & 15.6\\
\hline
\rowcolor{gray!20} Ours-Full  & \color{green}\ding{51} & 1200  & CEE+COMP & OE    & 226.3 & \color{green}\ding{51} & 441.3 & 8.1\\
\hline
\end{tabular}
\end{adjustbox}
\label{tab:0}
\vspace{-15pt}
\end{table*}
\section{Benchmark}
\subsection{Collection}
\label{sec:3-1}
We describe our comprehensive data collection process that spans five key stages: \textbf{Acquisition}, \textbf{Standardization}, \textbf{Translation}, \textbf{Search Prevention}, and \textbf{Difficulty Classification}.
\par
\textbf{Acquisition:}
We collect public physics problems from global college entrance examinations, their associated practice tests, and international physics competitions. 
Our sources include Chinese, Indian, and Russian exams, as well as IPhO, APhO, EPhO, and so on.
This comprehensive benchmark derives from 1,254 PDFs containing over 20,000 unique problems, ensuring diverse difficulty levels.
\par
\textbf{Standardization:}
Using MinerU \cite{wang2024mineruopensourcesolutionprecise} framework, we parse the content of these PDFs into structured problem information.
Subsequently, all problems undergo rigorous deduplication, filtering, and formatting to ensure complete problem statements, precise physics terms, accurate expressions, and consistent presentation style.
\par
\textbf{Translation:} 
We implement a two-phase translation process utilizing translators for initial conversion.
Engineering Ph.D. candidates with physics expertise then verify the translations for accuracy and professionalism, especially physics terms.
\par
\textbf{Search Prevention:} 
Following \cite{rein2024gpqa}, we exclude problems whose solutions and answers can be found through a five-minute Google search to minimize potential data leakage.
\par
\textbf{Difficulty Classification:} 
Based on the time students typically need to solve problems and the theorems applied, questions are categorized into knowledge-based and reasoning-based types, with the latter subdivided into three difficulty levels (easy, medium, and hard). 
This classification enables the comprehension evaluation of physics concepts and physics-based reasoning capabilities.
\subsection{Annotation}
\label{sec:3-2}
As shown in Figure \ref{fig:0}, our annotation framework consists of 8 key elements: \textbf{Diagram}, \textbf{Context}, \textbf{Sub-questions}, \textbf{Solution}, \textbf{Step Analysis}, \textbf{Answer}, \textbf{Theorem}, and \textbf{Difficulty}.
\textbf{Context} presents the physics scenario and conditions.
\textbf{Diagram} visualizes the physics scenario with concise illustrations complementing the \textbf{Context}.
\textbf{Sub-questions} give questions to assess the understanding and application of the concept.
\textbf{Solution} provides a step-by-step reasoning process, and \textbf{Answer} gives the answer to each sub-question.
\textbf{Step Analysis} explains the physics theorem used in each step and the physics quantities obtained.
\textbf{Theorem} lists the physics theorems applied in the question, and \textbf{Difficulty} indicates the difficulty classification.
\begin{figure*}[t]
\centering
\includegraphics[width=0.98\textwidth]{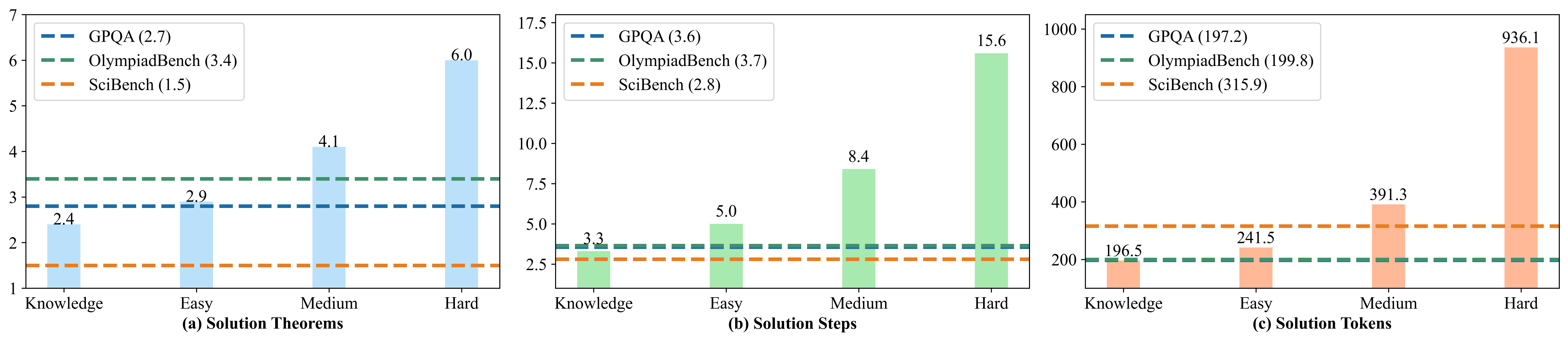}
\vspace{-10pt}
\caption{Analysis of solution theorems, solution steps, and solution tokens across different problem categories,  with comparisons from SciBench, GPQA, and OlympiadBench.}
\label{fig:11}
\vspace{-15pt}
\end{figure*}
\subsection{Characteristics}
\label{sec:3-3}
PhysReason consists of 1,200 carefully curated physics problems as shown in Table \ref{tab:0}, with a strategic composition of 25\% knowledge-based and 75\% reasoning-based questions across various difficulty levels, collectively covering 147 physics theorems.
The problems span Classic Mechanics, Quantum Mechanics, Fluid Mechanics, Thermodynamics, Electromagnetics, Optics, and Relativity.
As shown in Figure \ref{fig:11}, three critical solution metrics (theorem, step, and token) correlate positively with problem difficulty levels, validating the rationality of our difficulty classification.
Notably, the medium and hard problems demonstrate higher complexity compared to existing benchmarks.
PhysReason distinguishes itself through three key characteristics:
\begin{enumerate}[leftmargin=*,itemsep=0pt,parsep=0pt,topsep=0pt]
\item \textbf{Stratified difficulty}: 
The benchmark maintains a carefully balanced composition of knowledge-based (25\%) and reasoning-based (75\%) problems. 
The reasoning-based problems are methodically distributed across three difficulty levels (easy, medium, and hard - 25\% each), enabling comprehensive capability evaluation.
\item \textbf{Complex reasoning}: 
Detailed step-by-step solution annotations accompany each problem.
These annotated solutions demonstrate complex reasoning chains averaging 8.1 steps per problem, with hard problems requiring up to 15.6 steps, significantly surpassing the complexity of existing physics-based reasoning benchmarks.
\item \textbf{Multi-modal design}: The benchmark features a high proportion (81\%) of problems with  diagrams, authentically replicating physics-based reasoning scenarios while effectively evaluating both textual and visual reasoning capabilities.
\end{enumerate}
\begin{algorithm*}
\small
\caption{Physics Solution Auto Scoring Framework-Step Level (PSAS-S)}
\begin{algorithmic}[1]
\State \textbf{Phase 1: Data Extraction} \Comment{Extract and normalize solution steps from model output}
    \State \textit{Input}: Model output $M$, Annotation solution steps $S = \{s_1, s_2, ..., s_n\}$, Annotation step formulas $F = \{f_1, f_2, ..., f_n\}$,
    Annotation step values $V = \{v_1, v_2, ..., v_n\}$ \Comment{Information needed for step-level evaluation}
    \For{$s_i \in S$}
        \State $E_i \gets\text{LLM}(\text{ExtractTemplate}(M, s_i))$ \Comment{E: extracted relevant steps}
    \EndFor
    \State Assert $|E| = |S|$ \Comment{Ensure one-to-one mapping between extracted and annotated steps}
\State \textbf{Phase 2: Scoring} \Comment{Evaluate formula application and numerical calculations}
    \For{$(e_i, s_i) \in (E, S)$}
        \State $\hat{f}_i \gets \text{ExtractFormula}(e_i)$ \Comment{Formula content}
        \State $\hat{v}_i \gets \text{ExtractValue}(e_i)$ \Comment{Calculation target}
        \State ${score}_i \gets 0.5 \times \text{ScoreFormula}(\hat{f}_i, f_i) + 0.5 \times \text{ScoreValue}(\hat{v}_i, v_i)$ \Comment{Get final score}
    \EndFor
\State ${final\_score} \gets \frac{\sum_{i=1}^n{{{score}_i}}}{n}$ \Comment{Get the final score with the step-level evaluation}
\State \textbf{Phase 3: First Error Step Detection} \Comment{Identify the earliest point of solution deviation}
    \State $first\_error\_step \gets \infty$ \Comment{Initialization}
    \For{$i \gets 1$ to $|S|$}
        \If{${score}_i < 1$}
            \State $error\_step \gets \text{FindOriStep}(M, e_i)$ with the relationship between $E$ and $M$\Comment{Find corresponding original step}
            \State $first\_error\_step \gets \text{min}(first\_error\_step, error\_step)$ \Comment{Get the minimum}
        \EndIf
    \EndFor
\State \textbf{Phase 4: Error Analysis} \Comment{Analyze the first error step}
    \State $\text{ErrorTypes } \mathcal{T} = \{\text{DAE}, \text{PTAE}, \text{PCAE}, \text{PPUE}, \text{VRE}, \text{CPE}, \text{BCAE} \}$ \Comment{Error categories}
    \If{$first\_error\_step < \infty$}
    \State $j \gets first\_error\_step$
        \State $error\_type \gets \text{LLM}(\text{ClassificationTemplate}(e_{j}, s_{j}, \mathcal{T}))$ \Comment{Identify error type}
        \State $error\_analysis \gets \text{LLM}(\text{AnalysisTemplate}(e_{j}, s_{j}))$ \Comment{Generate error analysis}
    \EndIf
    \State \textit{Output}: ${final\_score}$, $first\_error\_step$, $error\_type$, $error\_analysis$
\end{algorithmic}
\label{alg:scoring}
\end{algorithm*}
\section{Evaluation Framework}
\subsection{Why LLMs Can Evaluate?}
Unlike multiple-choice problems, PhysReason contains open-ended answers and steps with diverse expressions but consistent semantics.
Given that LLMs have demonstrated exceptional capabilities in both precise content extraction and formula consistency evaluation \cite{2023opencompass, gao2024omni}, they serve as practical tools for automated physics solution evaluation.
Therefore, we propose automated answer-level and step-level evaluations, achieving comprehensive evaluation and avoiding labor-intensive manual assessment.
\subsection{How Answer-Level Evaluation Works?}
\label{Answer-Level Evaluation}
We develop Physics Solution Auto Scoring Framework-Answer Level (PSAS-A), which evaluates based on sub-question answers.
Given a model's reasoning process $M$ for a problem with sub-questions $\{q_1, q_2, \ldots, q_n\}$, we first extract the model's answers $\hat{a}_{i}$ for each $q_{i}$ from $M$ with an LLM.
Then, we employ the LLM to verify if $\hat{a}_{i}$ is semantically consistent with the standard answer $a_{i}$ of sub-question $q_i$.
The comparison function $C(\hat{a}_{i}, a_{i})$ returns 1 if consistent and 0 otherwise.
Considering that the sub-questions with different steps should not carry equal weights in scoring, we use the length of annotation solution $s_i$ of sub-question $q_i$, i.e., $|(s_i)|$ as a weighting scalar.
The model's reasoning process $M$'s answer-level score for each problem is calculated as follows:
\begin{equation}
\text{Score}(M) = \frac{\sum_{q_i} |(s_i)| \times C(\hat{a}_{i}, a_{i})}{\sum_{q_i} |(s_i)|}
\end{equation}
\subsection{What is the Step?}
Considering the correct execution of the reasoning process, each step should satisfy the following three conditions:
\textbf{Completeness}: Each step should contain a complete unit of logical reasoning
\textbf{Independence}: Each step should be understandable and evaluable as an independent logical unit
\textbf{Progression}: Each step should provide substantial progress in the problem-solving process, moving the solution forward
Therefore, we define that each step in the annotation must include a formula derived from applying a physics-based theorem and its related calculations.
Through physics theorems and formulas, we ensure that our defined steps maintain completeness, independence, and progression in physical reasoning, providing a solid foundation for subsequent content.
\subsection{How Step-Level Evaluation Works?}
\label{Step-Level Evaluation}
The current mainstream evaluation approach \cite{he-etal-2024-olympiadbench} with LLMs relies on answers, failing to reveal how and where models deviate from correct reasoning paths. 
To address this, we propose the Physics Solution Auto-Scoring Framework-Step Level (PSAS-S), which enables detailed assessment and analysis of each reasoning step.
The framework is divided into four phases: \textbf{Data Extraction}, \textbf{Scoring}, \textbf{First Error Step Detection}, \textbf{Error Analysis}, as detailed in Algorithm~\ref{alg:scoring}.
\par
\textbf{Data Extraction} phase leverages LLMs using \textit{Target} components from \textit{Step Analysis} annotations (Figure \ref{fig:0}) as prompts to locate and extract relevant content from model outputs for each annotated solution step $s_i$.
This phase effectively handles redundant thinking processes in LLM's reasoning process while maintaining semantic equivalence.
It obtains the mapping relationship between extracted relevant steps $E$ and annotated solution steps $S$.
\par
\textbf{Scoring} phase evaluates each step $s_i$ through two complementary components of theorem assessment $\textit{ScoreFormula}(\hat{f}_i, f_i)$ and result verification $\textit{ScoreValue}(\hat{v}_i, v_i)$, each with a weight of 0.5.
The final score is calculated as shown in Algorithm~\ref{alg:scoring}.
This ensures a balanced assessment of theorem application and computational accuracy.
\par
\begin{figure}[t]
\vspace{-10pt}
\centering
\includegraphics[width=0.48\textwidth]{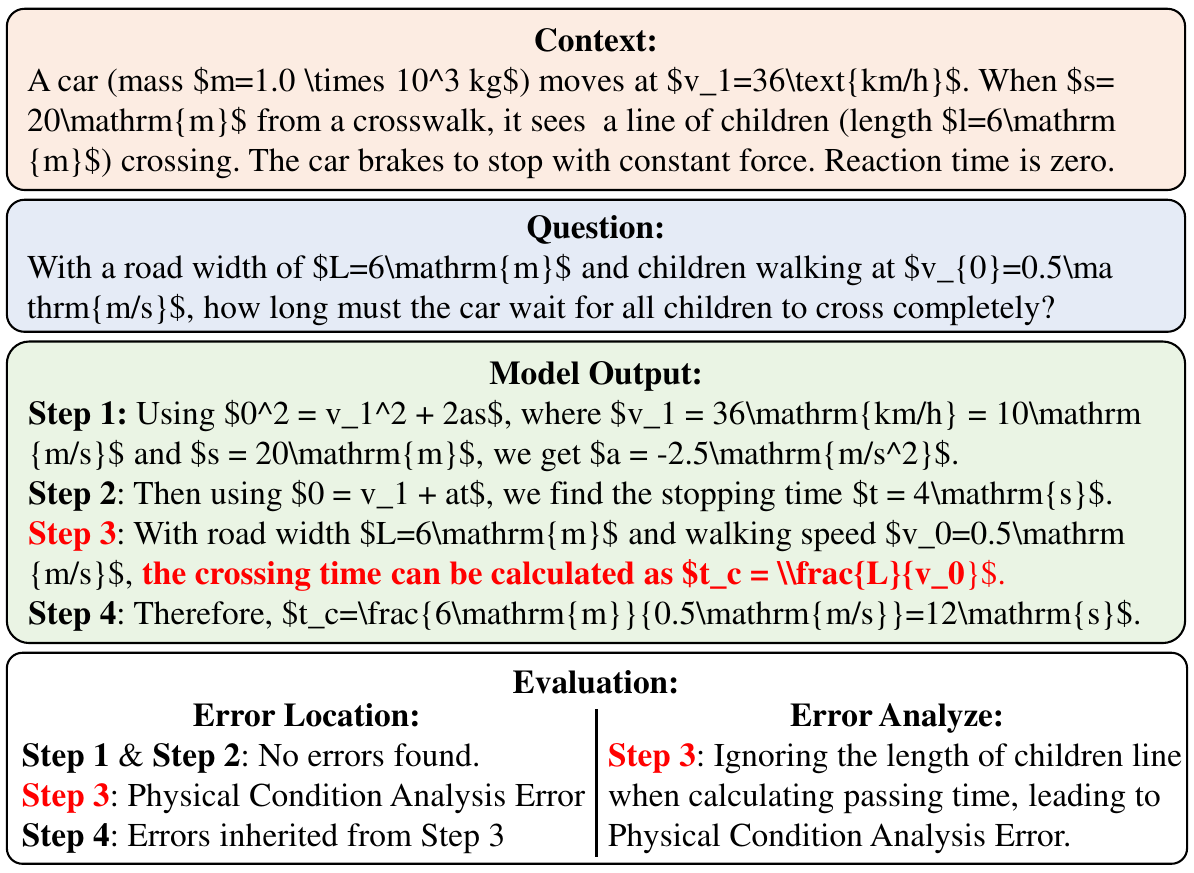}
\vspace{-20pt}
\caption{Step-level evaluation example obtained from PSAS-S framework.}
\label{fig:20}
\vspace{-20pt}
\end{figure}
\par
\textbf{First Error Step Detection} phase identifies the earliest step of deviation from the correct solution path.
When any step is found with a score below $1$, \text{FindOriStep} function locates the corresponding original step in the model's raw output based on the mapping relationship between $E$ and $S$ obtained from the \textbf{Data Extraction} phase, and updates $first\_error\_step$ to maintain the earliest error position.
This enables precise identification of where the model's reasoning first goes wrong. 
\par
\textbf{Error Analysis} phase analyzes the first error step detected in the solution, with two components: error classification and error analysis. 
For error classification, PSAS-S considers seven types of common errors: Diagram Analysis Error (DAE), Physics Theorem Application Error (PTAE), Physics Condition Analysis Error (PCAE), Physics Process Understanding Error (PPUE), Variable Relationship Error (VRE), Calculation Process Error (CPE), and Boundary Condition Analysis Error (BCAE). 
Detailed error-type descriptions are available in the Appendix \ref{Error Type Details}.
LLMs use structured prompts to identify the error type for the first error step. 
Then, a comprehensive error analysis is generated to explain the reasoning behind the mistake. 
A simplified example is shown in Figure \ref{fig:20}.
\begin{table}[t]
\vspace{-10pt}
\caption{Comparison between PSAS framework and direct use of LLM evaluation, where Answer Acc. denotes the accuracy of answer-level evaluation and Step Acc. indicates the precision in identifying the initial error step in the reasoning process.}
\vspace{-5pt}
\begin{adjustbox}{width=\columnwidth}
\centering
\begin{tabular}{lcc}
\hline
Model & Answer Acc. & Step Acc. \\
\hline
Gemini-2.0-Flash & 87.81 & 33.18 \\
Deepseek-V3 & 89.78 & 34.45\\
Gemini-2.0-Flash-Thinking-0121 & 91.24 & 35.74 \\
Deepseek-R1 & 93.31 & 37.54\\
\hline
Our (Gemini-2.0-Flash) & 98.96 & 97.23 \\
Our (Deepseek-V3) & 99.35 & 98.04 \\
\hline
\end{tabular}
\end{adjustbox}
\label{tab:5}
\vspace{-15pt}
\end{table}
\subsection{Whether Evaluation Trustworthy?}
\label{Framework Accuracy Analysis}
To validate the reliability of both our PSAS-A and PSAS-S, we compare our PSAS against conventional direct LLM evaluation approaches at both answer-level and step-level, using the Chain-of-Thought reasoning strategy.
We implement experiments using Deepseek-V3 and Gemini-2.0-Flash as scoring models in the following experiments:
\begin{enumerate}[leftmargin=*,itemsep=0pt,parsep=0pt,topsep=0pt]
\item 
For answer-level evaluation, we employ scoring models to assess answer correctness by combining both model-generated outputs and annotation answers.
We then compare these results with the judgments obtained from PSAS-A.
\item For step-level evaluation, inspired by previous work \cite{zheng2024processbench}, we design the task of identifying the first error step in reasoning processes containing errors, where higher accuracy indicates a more precise evaluation of the reasoning process.
Then, we submit both model-generated and annotated reasoning processes to scoring models to determine the location of the first error step, comparing with PSAS-S. 
\end{enumerate}
\par
Then, we collect 8,400 reasoning processes generated from multiple advanced models, including {Deepseek-R1}, {Gemini-2.0-Flash}, {Gemini-2.0-Flash-Thinking-0121}, {GLM-Zero}, {o1-mini}, {QwQ-32B}, and {QvQ-72B}.
Subsequently, we randomly sample 1,000 reasoning processes and meticulously manually annotate them to determine the correctness of each answer and identify the location of the first error step.
The results presented in Table \ref{tab:5} demonstrate that our frameworks achieve superior performance compared to direct LLM evaluation, highlighting the accuracy and reliability of PSAS evaluation results on PhysReason.

\par
\par
\section{Experiments}
\begin{table*}[t]
\centering
\caption{Model performance comparisons on the PhysReason benchmark using answer-level (left of /) and step-level (right of /) evaluations across different input combinations of Questions (Q), Images (I), and Image Captions (IC).
Gemini-2.0-T$^{\dagger}$ and $^{*}$ represent Gemini-2.0-Flash-Thinking-1206 and 0121.}
\vspace{-5pt}
\begin{adjustbox}{width=\textwidth}
\begin{tabular}{lcccccc}
\hline
Model & Input & Knowledge & Easy & Medium & Hard & Avg. \\
\hline
\rowcolor{gray!20} \multicolumn{7}{c}{\textbf{Non-O-like Models}} \\
Qwen2VL-72B & Q, I & 41.92/62.47 & 24.04/45.26 & 15.97/36.13 & 4.83/24.23 & 16.96/42.88\\
InternVL2.5-78B & Q, I & 28.34/64.71 & 24.16/50.69 & 17.72/38.56 & 9.71/25.95 & 19.98/45.89\\
GPT-4o & Q, I & 50.71/65.82 & 33.87/51.98 & 22.73/42.36 & 11.03/24.71 & 29.58/47.23\\
Deepseek-V3-671B & Q, IC & 55.86/66.14 & 40.06/52.77 & 26.63/44.02 & 13.73/26.87 & 34.07/48.42\\
Claude-3.5-Sonnet & Q, I & 54.14/66.45 & 41.35/55.85 & 28.14/44.86 & 15.11/28.51 & 34.69/49.88\\
Gemini-2.0-Flash & Q, I & 65.08/75.04 & 54.84/68.60 & 39.79/55.67 & 21.99/38.39 & 45.20/60.40 \\
Gemini-2.0-Pro & Q, I & 67.99/79.01 & 55.43/71.47 & 44.29/57.74 & 23.81/42.66 & 47.88/62.74 \\
\hline
\rowcolor{gray!20} \multicolumn{7}{c}{\textbf{O-like Models}} \\
o1-mini & Q, IC & 53.90/65.74 & 35.21/52.26 & 22.24/40.19 & 10.61/26.80 & 30.49/47.18\\
QvQ-72B & Q, I & 62.44/70.92 & 53.74/64.65 & 28.18/54.88 & 14.30/36.47 & 32.67/57.66\\
Gemini-2.0-T$^{\dagger}$ & Q, I & 65.35/77.20 & 51.89/67.49 & 44.43/58.95 & 27.14/45.48 & 47.20/63.07 \\
QwQ-32B & Q, IC & 62.03/76.28 & 54.92/71.08 & 43.64/62.14 & 22.99/42.19 & 45.89/63.87 \\
GLM-Zero & Q, IC & 64.95/80.36 & 54.11/71.54 & 41.32/63.67 & 23.04/47.46 & 46.52/65.76\\
o3-mini-high & Q, IC & 70.67/83.61 & 67.20/81.95 & 45.31/64.57 & 30.12/47.23 & 53.32/69.34\\
Gemini-2.0-T$^{*}$ & Q, I & 73.44/84.15 & 63.17/75.94 & 50.41/66.60 & 31.90/48.47 & 54.73/69.73 \\
Deepseek-R1 & Q, IC & 75.11/85.91 & 65.08/79.81 & 54.84/72.02 & 31.95/51.50 & 56.75/73.26 \\
\hline
\end{tabular}
\end{adjustbox}
\label{tab:3}
\vspace{-10pt}
\end{table*}
\begin{table}[t]
\caption{Comparison on PhysReason-mini with PSAS-A, where Gemini-2.0-T$^{\dagger}$ and $^{*}$ represent Gemini-2.0-Flash-Thinking-1206 and 0121. And K., E., M. and H. represent knowledge, easy, medium and hard.}
\vspace{-5pt}
\begin{adjustbox}{width=\columnwidth}
\begin{tabular}{lccccc}
\hline
Model & K. & E. & M. & H. & Avg. \\
\hline
o1-mini & 54.80 & 30.33 & 15.41 & 7.92 & 27.11 \\
QvQ-72B & 51.17 & 37.10 & 29.83 & 22.13 & 35.06 \\
QwQ-32B & 64.40 & 50.07 & 38.88 & 27.45 & 45.20 \\
Gemini-2.0-T$^{\dagger}$ & 71.47 & 49.97 & 36.83 & 22.97 & 45.42 \\
GLM-Zero & 72.70 & 50.17 & 43.42 & 24.70 & 47.75 \\
o1 & 72.47 & 53.37 & 49.31 & 25.32 & 50.12 \\
o3-mini-high & 71.10 & 63.20 & 47.02 & 31.93 & 53.31\\
Gemini-2.0-T$^{*}$ & 76.33 & 56.87 & 51.85 & 32.61 & 54.42 \\
Deepseek-R1 & 85.17 & 60.77 & 47.24 & 33.23 & 56.60 \\
\hline
\end{tabular}
\end{adjustbox}
\label{tab:4}
\vspace{-20pt}
\end{table}
\subsection{Setting}
\textbf{Baselines:}
We evaluate current mainstream open-source and closed-source LLMs, VLMs, and several o-like models. 
For models that cannot accept visual inputs, we use Gemini-2.0-Flash to generate captions for each image as supplementary information.
We assess 15 advanced LLMs/VLMs under the zero-shot Chain-of-Thought (CoT) setting (encouraging models to think step by step), including 7 non-O-like models (Qwen2-VL-72B \cite{wang2024qwen2}, GPT-4o \cite{openai_gpt4}, Claude-3.5-Sonnet \cite{anthropic_claude}, InternVL2.5-78B \cite{chen2024expanding}, Deepseek-v3 \cite{deepseekai2024deepseekv3technicalreport}), Gemini-2.0-Flash \cite{google_gemini2}, Gemini-2.0-Pro \cite{google_gemini2_pro} and 8 O-like models  (QwQ-32B \cite{qwq-32b-preview}, QvQ-72B \cite{qvq-72b-preview}, o1-mini \cite{o1_mini}, o1 \cite{openai2024learning}, o3-mini-high \cite{o3_mini}, Gemini-2.0-Flash-Thinking \cite{google_gemini_thinking}, Deepseek-R1 \cite{guo2025deepseek}, GLM-Zero \cite{glm_zero}).
Note that Gemini-2.0-Flash-Thinking has two versions: 1206 and 0121.
Due to API limitations, we do not experiment with o1 on the entire dataset. 
All other models are evaluated on the complete benchmark.
\par
\textbf{Evaluation Workflow:}
We encourage models to generate reasoning processes step by step for all problems in PhysReason, with open-source models running on NVIDIA A800 GPUs.
Please refer to Appendix-\ref{evaluation_prompt} for the detail prompt template.
Then, we evaluate the models' performance with the PSAS framework at both the answer and step levels, as described in Sections \ref{Answer-Level Evaluation} and \ref{Step-Level Evaluation}.
Based on the experimental results in Section \ref{Framework Accuracy Analysis}, considering both efficiency and performance, we select Deepseek-V3 as the final scoring model.
\par
\textbf{PhysReason-mini:}
Considering that the complete PhysReason requires relative high evaluation costs, we create a balanced PhysReason subset - PhysReason-mini.
We randomly sample 200 questions from the whole benchmark (50 for each difficulty level), striving to achieve equal representation across categories wherever possible.
\subsection{Main Results}
As demonstrated in Tables \ref{tab:3} and \ref{tab:4}, the experimental results on the PhysReason and PhysReason-mini reveal several significant findings.
\par
\textbf{Model Categories:}
O-like models exceed non-O-like ones, with multiple O-like models surpassing 50\% answer-level accuracy compared to non-O-like models' peak of 47.88\%.
\par
\textbf{Difficulty Level Analysis:}
As the difficulty increases, the required solution steps also increase, while model performance severely declines, indicating that models still perform inadequately on physics problems requiring deep reasoning.
\par
\textbf{Step-level vs. Answer-level Evaluation:}
The two evaluation frameworks assess performance from different perspectives.
Step-level scores consistently surpass answer-level scores, indicating that models can achieve some correct steps despite failing to reach the correct final answer.
Moreover, the step-level score differences between models become more pronounced than those at the answer level as problem difficulty increases.
This demonstrates that step-level evaluation proves more discriminative in distinguishing model capabilities, particularly in highly challenging problems.
The distributions of these two evaluation methodologies exhibit non-perfect synchronization, indicating that step-level evaluation provides comprehensive insights to answer-level assessment.
\par
\textbf{Medium and Hard Problem Analysis:}
Performance on medium and hard reasoning problems can emerge as key differentiators of model physics-based reasoning ability.
Among these models, those achieving scores of 40/60 and 30/50 on answer-level and step-level evaluations respectively serve as critical reference points.
\par
\textbf{Knowledge-Reasoning Correlation Analysis:}
Results show a positive correlation between physics knowledge and reasoning capabilities, with Deepseek-R1 and Gemini-2.0-Flash-Thinking-0121 excelling in both aspects. 
Moreover, among models with similar scores on knowledge problems, O-like models tend to achieve higher scores on reasoning problems (as demonstrated by Gemini-2.0-Flash and Gemini-2.0-T$^{\dagger}$).
This suggests that reinforcement learning and training with thought chains help improve models' reasoning capabilities.
In conclusion, effective reasoning relies on knowledge capacity and model architecture.
\begin{table}[t]
\caption{Test-Time Compute Scaling Performance Comparisons on PhysReason-mini with PSAS-A, where Flash and Think denote Gemini-2.0-Flash and Gemini-2.0-Flash-Thinking-0121, and Tour. means Tournament.}
\vspace{-5pt}
\begin{adjustbox}{width=\columnwidth}
\centering
\begin{tabular}{lllcccc}
\toprule
{Base} & {Method} & {Reward} & N=1 & N=2 & N=4 & N=8 \\
\midrule
\multirow{4}{*}{Flash} 
    & \multirow{2}{*}{BoN} & Flash & 46.52 & 46.67 & 47.12 & 47.81\\
    & & Think & 46.52 & 47.37 & 48.87 & 50.94 \\
    \cmidrule{2-7}
    & \multirow{2}{*}{Tour.} & Flash & 46.52  & 45.87 & 47.36 & 49.58\\
    & & Think & 46.52 & 47.51 & 52.11 & 53.06\\
\midrule
\multirow{2}{*}{Think} 
    & BoN & Think & 54.42 & 52.27 & 54.78 & 55.13\\
    & Tour. & Think & 54.42 & 55.60 & 56.26 & 56.57\\
\bottomrule
\end{tabular}
\end{adjustbox}
\label{tab:10}
\vspace{-15pt}
\end{table}
\subsection{Results with Test-Time Compute Scaling}
We evaluate Best-of-N (BoN) and Tournament-Style selection \cite{snell2025scaling, yang2024qwen2} test-time compute scaling methods on PhysReason-mini. 
Using Gemini-2.0-Flash and Gemini-2.0-Flash-Thinking-0121 as base models, we test different reward model configurations: when Flash serves as base model, both itself and Thinking-0121 are evaluated as reward models, while Thinking-0121 uses self-reward due to its superior reasoning. 
Both methods \cite{cobbe2021training, lightmanlet, son2024varco} select optimal responses from multiple Chain-of-Thought candidates (N = 1, 2, 4, 8), as shown in Table \ref{tab:10}. 
These scaling methods demonstrate the potential to enhance model performance through strategic response selection and process reward modeling.
\begin{table}[t]
\caption{Performance Comparison with PSAS-A after Directly Concatenation (D. Acc) and Guided Error Localization (G. Acc) on PhysReason-mini, where Acc. means the original performance of the model, Gemini-2.0-T$^{*}$ represents Gemini-2.0-Flash-Thinking-0121.}
\vspace{-5pt}
\begin{adjustbox}{width=\columnwidth}
\centering
\begin{tabular}{lccc}
\hline
Model & Acc. & D. Acc. & G. Acc.\\
\hline
Deepseek-V3 & 34.07 & 29.31 & 40.78\\
Gemini-2.0-Flash & 46.52 & 42.76 & 51.55\\
Gemini-2.0-T$^{*}$ & 54.42 & 50.66 & 56.82\\
Deepseek-R1 & 56.60 & 52.26 & 58.33\\
\hline
\end{tabular}
\end{adjustbox}
\label{tab:6}
\vspace{-10pt}
\end{table}
\subsection{Performance Improving with PSAS-S}
Given PSAS-S's capability to locate and analyze the first error step as presented in Section \ref{Step-Level Evaluation}, we conduct experiments on PhysReason-mini to explore whether models can correct errors after being informed.
The experiments are divided into \textit{Direct concatenation} and \textit{Guided error localization}.
The former (D. Acc.) combines questions with the previous reasoning process for a second attempt.
For the latter (G. Acc.), PSAS-S is used to locate and analyze the first error in the reasoning process, then combines the question, previous reasoning, and \textbf{the location and analysis of the first error} for a second attempt.
As shown in Table \ref{tab:6}, results show that direct concatenation decreased performance by 3-5\%, while guided error localization improved performance by 3-6\%.
This suggests that guiding LLMs to identify reasoning errors is crucial for enhancing their reasoning capabilities and also proves the effectiveness of our PSAS framework.
\begin{figure}[t]
    \centering
    \includegraphics[width=0.5\textwidth]{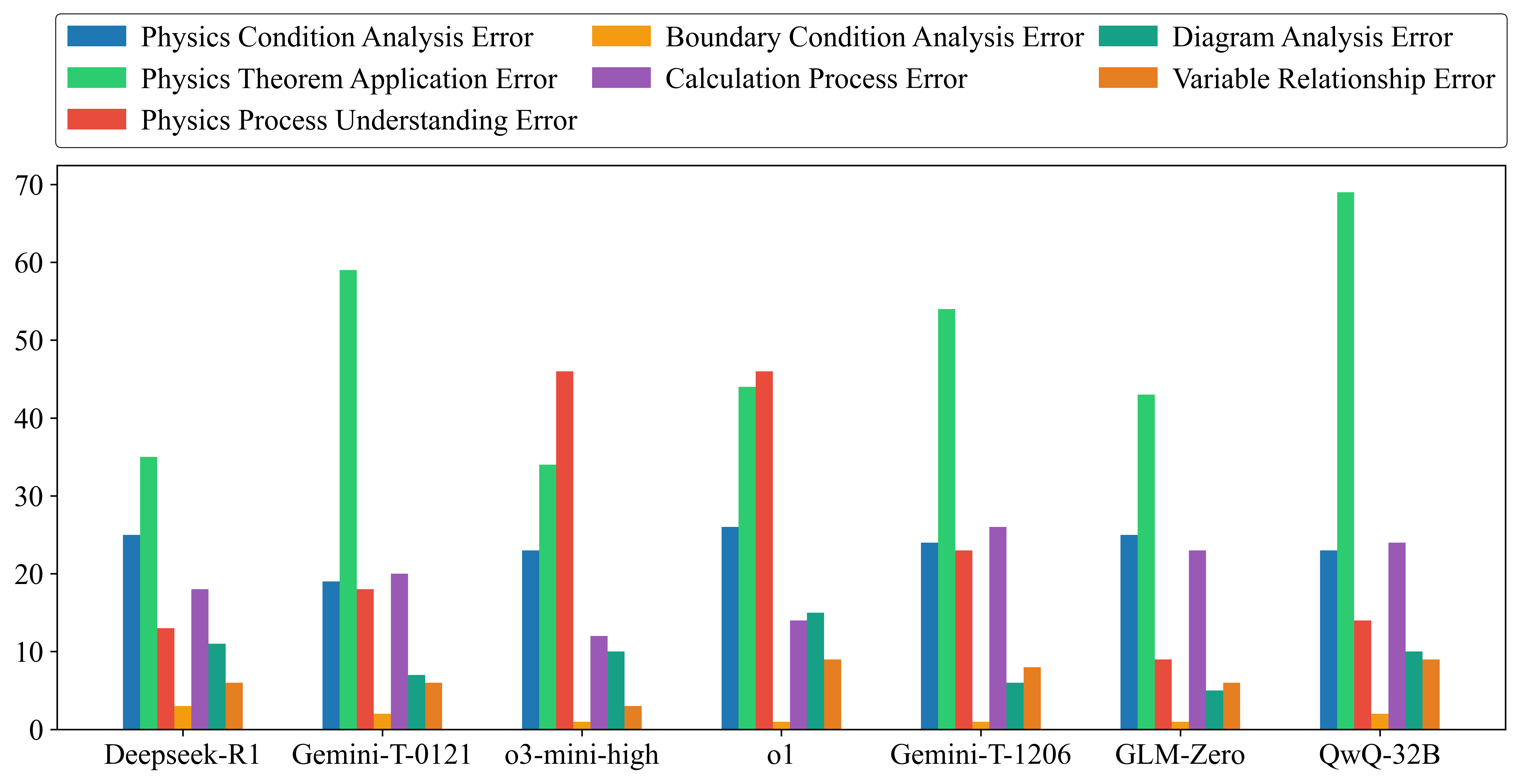}
    \vspace{-20pt}
    \caption{Error statistics with PSAS-S framwork in PhysReason-mini, where Gemini-T-1206 and Gemini-T-0121 denote Gemini-2.0-Flash-Thinking-1206 and Gemini-2.0-Flash-Thinking-0121. }
    \label{fig:3}
\vspace{-15pt}
\end{figure}
\subsection{Error Kind Distribution Analysis}
Discovering errors in reasoning processes is not equivalent to fully understanding them; it's also crucial to understand the causes of errors.
We analyze the error distributions of different models on PhysiReason-mini as shown in Figure \ref{fig:3}.
Four prevalent error types consistently challenge all models: Physics Theorem Application, Physics Process Understanding, Calculation Process, and Physics Condition Analysis.
This reveals models' limited intuitive physics understanding, highlighting the need for stronger physics-based reasoning capabilities.
Notably, o1 and o3-mini-high show elevated Physics Process Understanding Errors but reduced Calculation Process Errors.
This maybe suggest a trade-off between conceptual comprehension and computational precision.
\begin{figure}[t]
    \centering
    \includegraphics[width=0.5\textwidth]{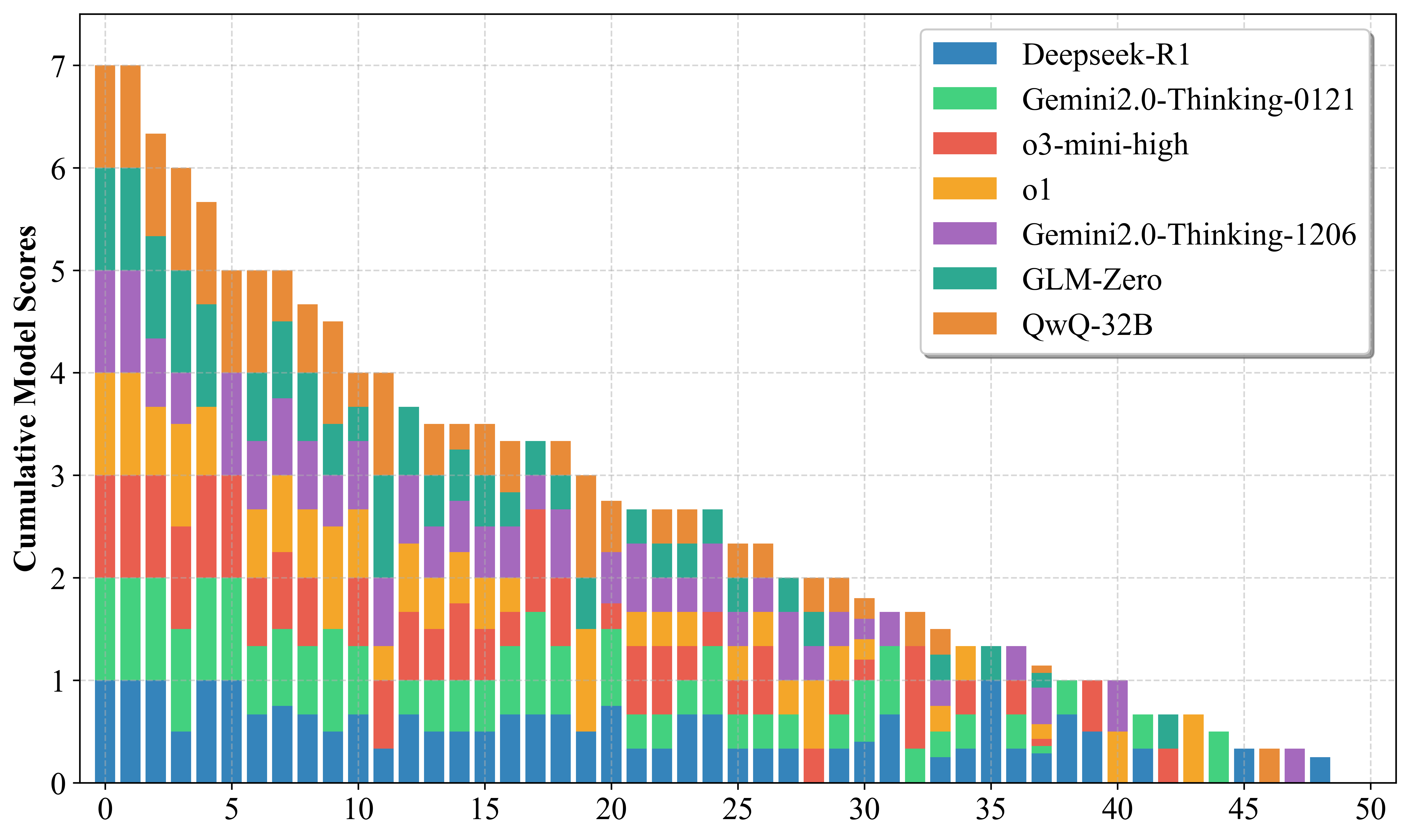}
    \vspace{-20pt}
    \caption{Performance with PSAS-S framework in the hard problems from PhysReason-mini.}
    \label{fig:4}
\vspace{-12pt}
\end{figure}
\subsection{Hard Problem Analysis}
Our analysis of 50 hard reasoning problems from PhysReason-mini across 7 models reveals two key insights (Figure \ref{fig:4}).
Despite variations in overall performance, each model exhibits unique strengths in specific problem domains, demonstrating the diverse nature of their reasoning capabilities.
The models' achievement of some scores (below $1$) is notable, indicating their ability to initiate correct solution paths but failing to maintain this accuracy throughout the reasoning process.
These patterns suggest that while current models grasp basic physics concepts, they struggle to sustain accurate reasoning across extended solution steps.
\section{Conclusion}
We introduce PhysReason, a novel physics-based reasoning benchmark with stratified difficulty and Physics Solution Auto-Scoring Framework with answer and step level evaluation. 
Experimental results show a consistent decline in performance as reasoning depth increases. 
This benchmark establishes new standards for evaluating and improving AI models' physics-based reasoning abilities.

\section{Acknowledgements}
This work was supported by the National Key Research and Development Program of China (2022YFC3303600), National Natural Science Foundation of China (No. 62137002, 62293550, 62293553, 62293554, 62450005, 62437002, 62477036, 62477037, 62176209, 62192781, 62306229),  `LENOVO-XJTU' Intelligent Industry Joint Laboratory Project, the Shaanxi Provincial Social Science Foundation Project (No. 2024P041), the Natural Science Basic Research Program of Shaanxi (No. 2023-JC-YB-593), the National Research Foundation, Singapore, under its NRF Fellowship (Award\# NRF-NRFF14-2022-0001), the Youth Innovation Team of Shaanxi Universities `Multi-modal Data Mining and Fusion', Project of China Knowledge Centre for Engineering Science and Technology, and the Youth AI Talents Fund of China Association of Automation (Grant No. HBRC-JKYZD-2024-311). Mike Shou does not receive any funding for this work.
\section*{Limitation}
Despite the comprehensive nature of our benchmark, two key limitations warrant discussion, concerning both benchmark construction and evaluation methodology.
First, they focus primarily on testing models' ability to apply and reason with physics theorems under idealized conditions, rather than fully reflecting real-world physics scenarios. 
However, it is worth noting that applying physics theorems under idealized conditions serves as the foundation for real-world physics scenarios, as the latter is more complex. 
However, current LLMs' performance even on idealized conditions remains unsatisfactory. 
Therefore, PhysReason remains valuable in evaluating models' ability to apply physics theorems for physics-based reasoning. 
Moreover, through data synthesis, many problems in PhysReason can be adapted to create real-world physics reasoning scenarios, which will be a direction for our future research.
Second, our evaluation framework, though achieving over 98\% accuracy using LLMs as assessment tools, is not without limitations. The PSAS-S framework, while demonstrating satisfactory performance, increases computational time for evaluation.
In future work, we will explore ways to optimize evaluation time while maintaining assessment accuracy.
\section*{Ethical Statement}
In developing PhysReason, we carefully considered and addressed potential implications and risks. 
Our benchmark, sourced exclusively from public official materials (IPhO, Gaokao, JEE, and authorized mock exams), undergoes rigorous data cleansing, deduplication, and standardization to ensure reliability while minimizing bias and data leakage. 
Committed to environmental sustainability, we publicly release complete datasets and accompanying scripts under appropriate licenses (MIT and CC BY-NC-SA) to cut down on unnecessary carbon footprint, while optimizing processing pipelines to reduce computational overhead. 
In all experiments, we strictly comply with all licenses for models and data. Our benchmark is an important resource that drives AGI's strength in scientific reasoning, maintaining high standards for data quality and ethical considerations.
\bibliography{custom}

\newpage
\appendix
\section{Data Sources}
Our dataset is derived from four distinct sources, each representing different levels and approaches to physics education and assessment. These sources have been carefully selected to ensure comprehensive coverage of physics problems across various difficulty levels and cultural contexts. The diversity of these sources helps in creating a robust and well-rounded dataset that captures different pedagogical approaches and problem-solving methodologies.
\begin{itemize}
\item \textbf{International Physics Olympiad (IPhO) Problems}\\
The Physics Olympiad problems are globally recognized for their complexity and quality. These problems typically require multiple solution approaches and the integration of capabilities across mathematics and physics sub-disciplines. 
Participants in these competitions represent some of the world's strongest talent in physics logical reasoning. 
The problems often combine theoretical understanding with practical applications, requiring students to demonstrate both analytical and creative problem-solving skills. The international nature of these competitions ensures a diverse range of problem-solving approaches and cultural perspectives.
\item \textbf{Chinese National College Entrance Examination (Gaokao) Physics Questions}\\
The Gaokao physics questions represent a rigorous standardized assessment system that has been refined over decades. These questions are designed to test both fundamental understanding and advanced application of physics concepts at the high school level. 
The problems are carefully calibrated to discriminate between different levels of student ability while maintaining high reliability and validity. 
They often incorporate real-world scenarios and practical applications, making them particularly valuable for assessing applied physics knowledge.
\item \textbf{Chinese Mock Examinations at Various Levels}\\
Our collection includes a comprehensive range of mock examination questions from multiple administrative levels in China. This includes provincial-level mock exams, city-level assessment materials, and joint examination papers created through collaboration among multiple high schools. 
These diverse sources provide a rich spectrum of problem-solving scenarios and difficulty levels. 
The multi-tiered nature of these mock examinations reflects different regional interpretations of educational standards while maintaining alignment with national requirements. The variety in question sources ensures exposure to different testing styles and pedagogical approaches, making this dataset particularly valuable for understanding the breadth of physics education assessment in China.
\item \textbf{Indian Joint Entrance Examination (Advanced)}\\
This examination represents one of India's most prestigious and challenging engineering entrance tests. The exam structure, consisting of two papers with 50-60 questions each, provides a comprehensive assessment of physics knowledge alongside mathematics and chemistry. 
The questions are known for their analytical depth and often require multi-step problem-solving approaches. 
The exam's high stakes nature and competitive environment ensure that the problems are both challenging and discriminating, making them valuable additions to our dataset.
\item  \textbf{Others}\\
We also obtained some physics questions from non-Chinese and English sources on hugging-face, such as Russian \cite{phy_big}.
This dataset consists of a diverse collection of physics problems, categorized into different domains, including 1000 problems on Kinematics, 600 problems on Electricity and Circuits, and 500 problems on Thermodynamics. All data has been extracted from open sources, ensuring a wide variety of problem types and difficulty levels.
\end{itemize}
The PhysReason benchmark is derived from publicly available physics education materials including:
International Physics Olympiad problems (2008-2021), Chinese National College Entrance Examination physics questions (2010-2024), Indian Joint Entrance Examination Advanced physics problems (2010-2024), Chinese provincial and municipal mock examination questions (2015-2024).
We have collected more than 20,000 physics problems.
All problems were collected in accordance with fair use principles for educational and research purposes. 
The complete benchmark and associated code will be released under the MIT License for research use.
\par
The dataset contains no personally identifiable information. 
All problems are from standardized tests and competition materials with no individual student data.
This documentation ensures reproducibility and proper usage of the benchmark while protecting privacy and intellectual property rights.
\begin{figure*}[t]
    \centering
    \includegraphics[width=\textwidth]{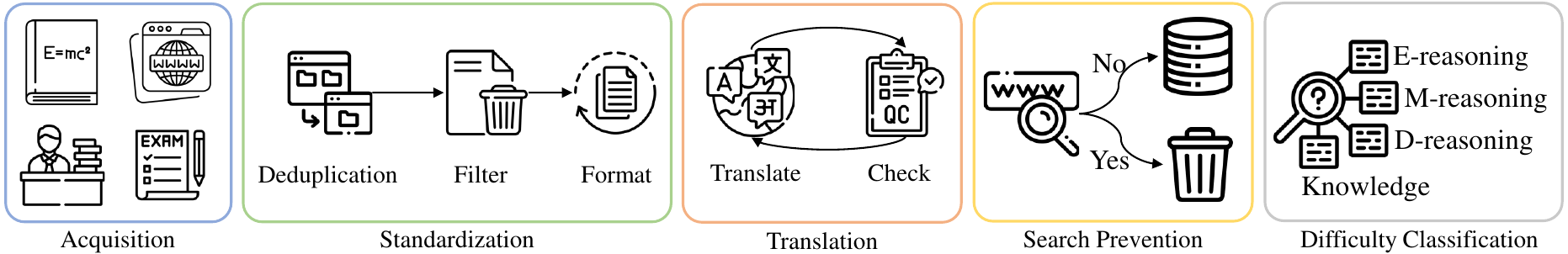}
    \caption{Illustration of the data collection pipeline.
}
    \label{fig:10}
\end{figure*}
\section{Benchmark}
\subsection{Collection}
\subsubsection{Data Acquisition}
We systematically collected, curated, and processed physics problems from diverse sources to ensure comprehensive coverage of physics concepts and problem-solving scenarios. 
Our dataset comprises 1,254 PDF documents totaling 27,874 pages, yielding over 20,000 unique problems. 
This extensive collection provides a rich foundation for developing a comprehensive physics problem benchmark.
\subsubsection{Data Standardization}
We implemented a systematic data processing pipeline utilizing MinerU \cite{wang2024mineruopensourcesolutionprecise} for PDF parsing. 
The standardization process encompasses several critical phases: initial format conversion, rigorous deduplication, and comprehensive formatting standardization.
Each question underwent a rigorous quality assessment process with specific evaluation criteria:
\begin{enumerate}
\item Complete problem statements with well-defined variables and conditions
\item Clear and unambiguous wording
\item Accurate expressions and units
\item Consistent formatting of equations and symbols
\end{enumerate}



\subsubsection{Translation}
To standardize the multilingual dataset comprising Chinese, English, Hindi, and Russian content, we implement a two-phase process:

\paragraph{Phase I: Translation}
\begin{itemize}
\item Initial translation by translators
\item Strict adherence to standardized physics terminology
\item Consistent mathematical notation and expressions
\end{itemize}

\paragraph{Phase II: Verification}
\begin{itemize}
\item Validation by Engineering Ph.D. candidates
\item Verification of physics terminology accuracy
\item Confirmation of semantic equivalence
\item Review of mathematical expression consistency
\end{itemize}

\subsubsection{Search Prevention}
Following \cite{rein2024gpqa}, we exclude problems whose answers could be found through a five-minute Google search to minimize data leakage. This step ensures that model evaluation reflects genuine physics problem-solving capabilities rather than information retrieval abilities.

\subsubsection{Difficulty Classification:}
Questions were systematically categorized using a multi-dimensional classification framework:
\paragraph{Primary Classification}
\begin{itemize}
\item  Knowledge-based questions:
    \begin{itemize}
    \item Focus on fundamental physics concepts
    \item Clear-cut application of specific theorems
    \item Direct calculation or concept identification
    \end{itemize}
\item Reasoning-based questions:
    \begin{itemize}
        \item Multiple-theorem integration
        \item Multi-step problem-solving approaches
        \item Complex analytical thinking
    \end{itemize}
\end{itemize}
\paragraph{Difficulty Levels in Reasoning-based Questions}
\begin{itemize}
\item Easy:
    \begin{itemize}
    \item Total steps $\leq$ 5
    \item Completion time: 0-5 minutes
    \end{itemize}
\end{itemize}

\begin{itemize}
\item Medium:
    \begin{itemize}
    \item Total steps $\leq$ 10
    \item Completion time: 5-15 minutes
    \end{itemize}
\end{itemize}

\begin{itemize}
\item Hard:
    \begin{itemize}
    \item Total steps $>$ 10
    \item Completion time: 15+ minutes
    \end{itemize}
\end{itemize}

\subsection{Annotation}
\paragraph{Key Elements}
As shown in Figure \ref{fig:0}, our annotation framework consists of 7 key elements:
\begin{itemize}
    \item Context:
    \begin{itemize}
        \item Detailed physics scenario description: Describe the physics setup thoroughly, including objects, environment, and interactions. For example, specify angles, materials, initial conditions, and forces.
        \item Clear specification of conditions and constraints: Explicitly list all given conditions: initial conditions (e.g., initial velocity, position), boundary conditions, and constraints (e.g., inextensible string, frictionless surface).
        \item Standardized notation for physics quantities: Use consistent and standard symbols for physics quantities (e.g., $v$ for velocity, $a$ for acceleration, $m$ for mass) throughout the annotation.
    \end{itemize}
\end{itemize}

\begin{itemize}
    \item Sub-question:
    \begin{itemize}
        \item Hierarchical structure of related questions: Break down a complex problem into smaller, logically connected sub-questions. These should build upon each other.
        \item Clear progression of complexity: Sub-questions should increase in difficulty, guiding the learner from basic concepts to more advanced analysis.
    \end{itemize}
\end{itemize}

\begin{itemize}
    \item Solution:
    \begin{itemize}
        \item Detailed step-by-step reasoning process: Provide a comprehensive, step-by-step solution.  Do not skip any crucial reasoning steps.
        \item Each step contains at least one formula:  Each step in the solution should include at least one relevant physics formula (theorem, law, or derived equation).
        \item If the formula can be solved to a value, it should also have a value: If a step's formula yields a numerical result, provide that result.
    \end{itemize}
\end{itemize}

\begin{itemize}
    \item Step Analysis:
    \begin{itemize}
       \item Explicit theorem application rationale: Clearly state which theorem, law, or principle is applied in each step and why it's applicable. Example: Newton's Second Law.
        \item Physics quantity derivation explanation: Explain how unknown physics quantities are derived from known ones. Example: "$W = \Delta E_k$"
    \end{itemize}
\end{itemize}

\begin{itemize}
    \item Answer:
    \begin{itemize}
        \item Numerical results with appropriate units: Provide the correct numerical value and units for numerical answers (e.g., "$v = 5 m/s$").
        \item Formulaic results with appropriate symbols: For formulaic answers, use previously defined standard symbols and ensure the formula's correctness (e.g., "$v = \sqrt{2gh}$").
    \end{itemize}
\end{itemize}

\begin{itemize}
    \item Difficulty:
    \begin{itemize}
        \item Reasoning difficulty metrics: Use qualitative descriptions (e.g., "knowledge" "easy," "medium," "hard") 
    \end{itemize}
\end{itemize}

\begin{itemize}
    \item Theorem:
    \begin{itemize}
        \item Comprehensive list of applicable theorems, laws, and formulas: Provide a complete list of all the specific physics theorems, laws, and equations that are relevant to solving the problem. Examples include: `Newton's Second Law', `Work-Energy Theorem', `Conservation of Momentum', `Kinematic Equations', etc.
        \item Core Concepts: Identify the fundamental physics principles and ideas that underpin the solution, even if they aren't expressed as a single equation. Examples include:  `Wave-Particle Duality'.
    \end{itemize}
\end{itemize}


\section{Error Type Details}
\label{Error Type Details}
The following is a summary of the error types, categorized and with expanded explanations:
\begin{enumerate}
    \item \textbf{Diagram Analysis Errors:}
    \begin{itemize}
        \item \textit{Description:} Errors related to the comprehension, plotting, analysis, or extraction of data from graphical representations. This encompasses any mistake made when working with diagrams, charts, or graphs.
        \item \textit{Examples:}
        \begin{itemize}
            \item Misreading the labels or units on the axes of a graph.
            \item Misinterpreting the trend of a curve (e.g., confusing a linear relationship with an exponential one).
            \item Failing to identify key data points or features on the graph (e.g., maxima, minima, intercepts).
            \item Incorrectly extrapolating or interpolating data from the graph.
            \item Drawing an inaccurate graph based on given data.
        \end{itemize}
    \end{itemize}

    \item \textbf{Physics Theorem Application Errors:}
    \begin{itemize}
        \item \textit{Description:} Errors arising from the incorrect application of physics theorems or principles, or using them in situations where they are not valid. This includes both misremembering the law itself and misapplying a correctly remembered law.
        \item \textit{Examples:}
        \begin{itemize}
            \item Applying Newton's Laws of Motion to a non-inertial reference frame without accounting for fictitious forces.
            \item Using the conservation of energy principle in a system where non-conservative forces (like friction) are doing significant work.
            \item Applying a formula outside of its valid range of applicability (e.g., using a small-angle approximation when the angle is large).
            \item Misunderstanding the conditions under which a particular law is valid.
        \end{itemize}
    \end{itemize}

    \item \textbf{Physics Condition Analysis Errors:}
    \begin{itemize}
        \item \textit{Description:} Errors related to the incorrect assessment of the physics system's boundaries, the forces acting on it, or its constituent components. This involves a misunderstanding of `what' is happening in the system.
        \item \textit{Examples:}
        \begin{itemize}
            \item Neglecting the force of friction in a situation where it is significant.
            \item Incorrectly identifying the system boundary, leading to errors in applying conservation laws.
            \item Misjudging whether a system is isolated (no external forces) or not.
            \item Failing to consider all relevant forces acting on an object.
            \item Misidentifying the components of a system that are interacting.
        \end{itemize}
    \end{itemize}

    \item \textbf{Physics Process Understanding Errors:}
    \begin{itemize}
        \item \textit{Description:} Errors stemming from a misunderstanding of how a physics phenomenon develops, how states change, or the causal relationships between events. This involves a misunderstanding of `how' things are happening.
        \item \textit{Examples:}
        \begin{itemize}
            \item Incorrectly analyzing the motion of a projectile, such as misunderstanding the independence of horizontal and vertical motion.
            \item Misunderstanding the mechanisms of energy transformation (e.g., confusing heat and temperature).
            \item Incorrectly predicting the direction of motion based on the forces involved.
            \item Having misconceptions about the nature of a physics process (e.g., believing that a continuous force is needed to maintain constant velocity).
        \end{itemize}
    \end{itemize}

    \item \textbf{Variable Relationship Errors:}
    \begin{itemize}
        \item \textit{Description:} Errors caused by misunderstanding the dependencies or functional relationships between different physics quantities. This involves incorrectly relating variables.
        \item \textit{Examples:}
        \begin{itemize}
            \item Incorrectly assuming that acceleration is directly proportional to velocity.
            \item Misunderstanding the relationship between force, mass, and acceleration (Newton's Second Law).
            \item Confusing the relationship between potential and kinetic energy.
            \item Failing to recognize an inverse relationship between two variables.
        \end{itemize}
    \end{itemize}

    \item \textbf{Calculation Process Errors:}
    \begin{itemize}
        \item \textit{Description:} Errors occurring during the mathematical manipulation of equations, the derivation of formulas, or the substitution of numerical values. These are purely mathematical mistakes.
        \item \textit{Examples:}
        \begin{itemize}
            \item Making algebraic errors when rearranging equations.
            \item Incorrectly performing unit conversions (e.g., mixing up meters and centimeters).
            \item Making arithmetic errors (e.g., simple addition or multiplication mistakes).
            \item Incorrectly substituting values into a formula.
            \item Errors in using a calculator.
        \end{itemize}
    \end{itemize}

    \item \textbf{Boundary Condition Analysis Errors:}
    \begin{itemize}
        \item \textit{Description:} Errors resulting from neglecting or mishandling special cases, limiting conditions, or the applicable ranges of variables or equations. This involves not considering the "edges" of the problem.
        \item \textit{Examples:}
        \begin{itemize}
            \item Failing to consider the behavior of a system at extremely high or low temperatures.
            \item Neglecting the effects of air resistance when analyzing projectile motion at high speeds.
            \item Not considering the limitations of a particular model or approximation.
            \item Applying a formula outside its range of validity.
            \item Ignoring initial conditions or other constraints.
        \end{itemize}
    \end{itemize}
\end{enumerate}
\begin{figure*}[t]
\centering
\includegraphics[width=\textwidth]{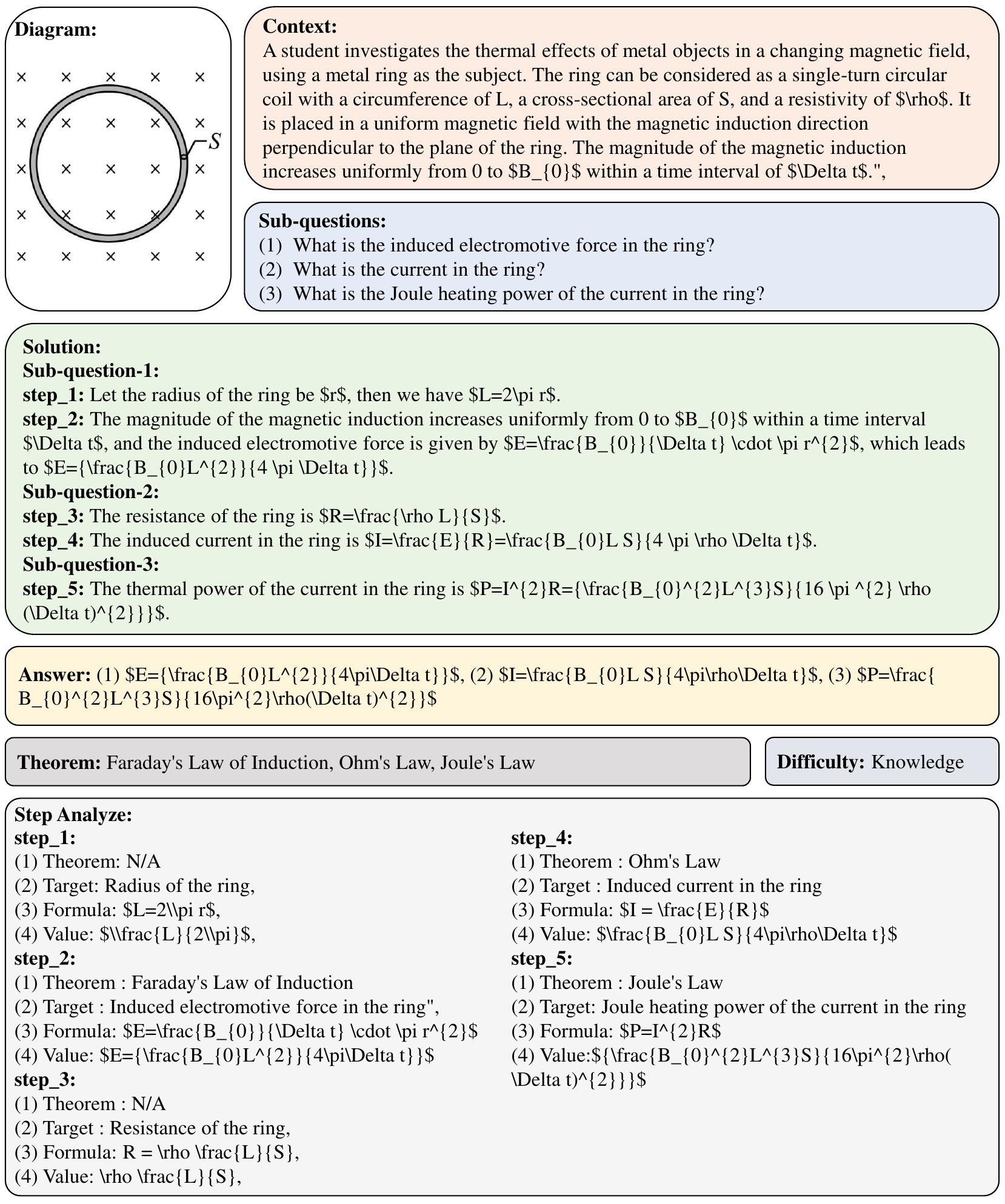}
\caption{A knowledge example in our benchmark.}
\label{fig:sup_1}
\end{figure*}

\begin{figure*}[t]
\centering
\includegraphics[width=\textwidth]{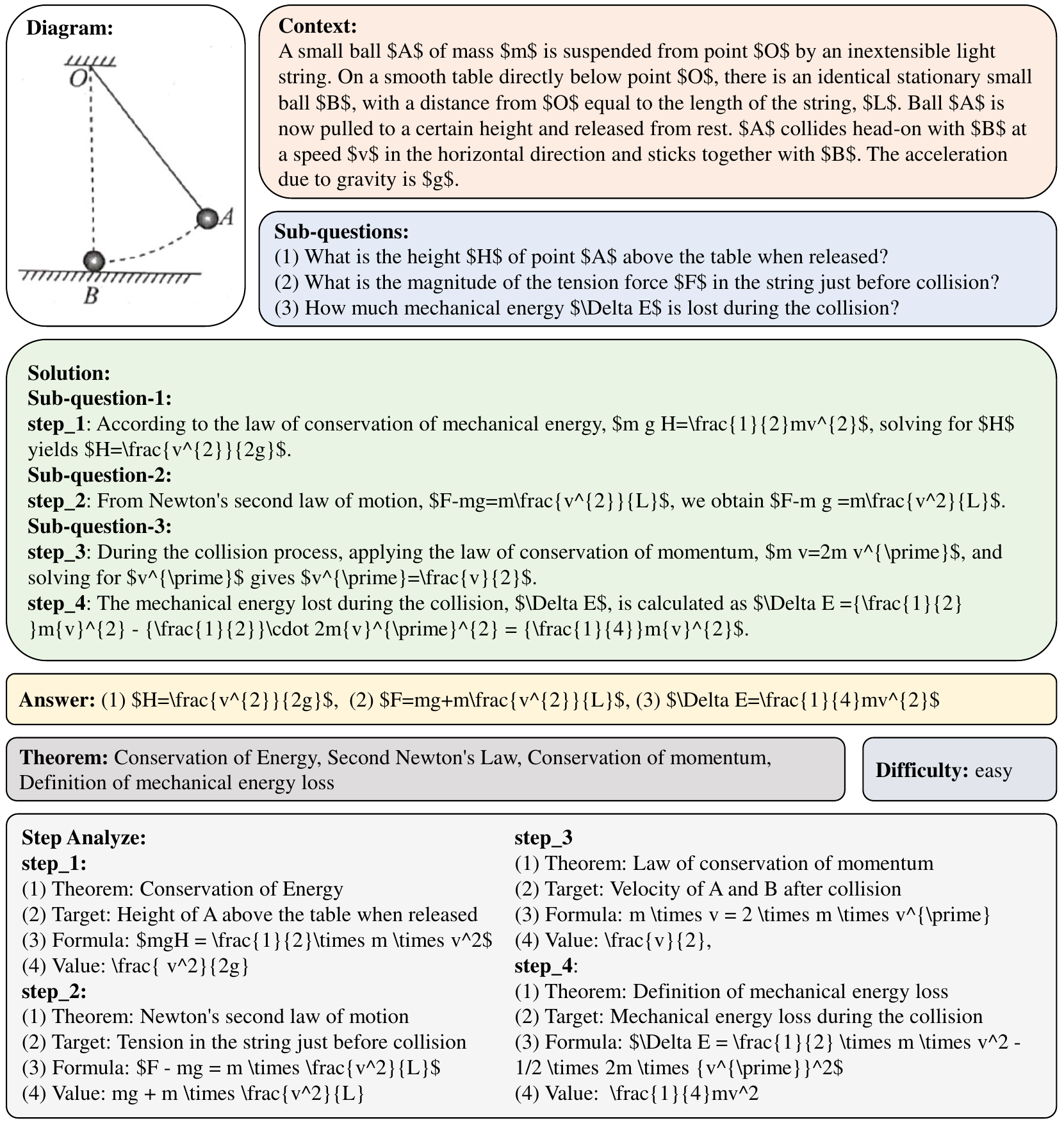}
\caption{An easy example in our benchmark.}
\label{fig:sup_0}
\end{figure*}

\begin{figure*}[t]
\centering
\includegraphics[width=\textwidth]{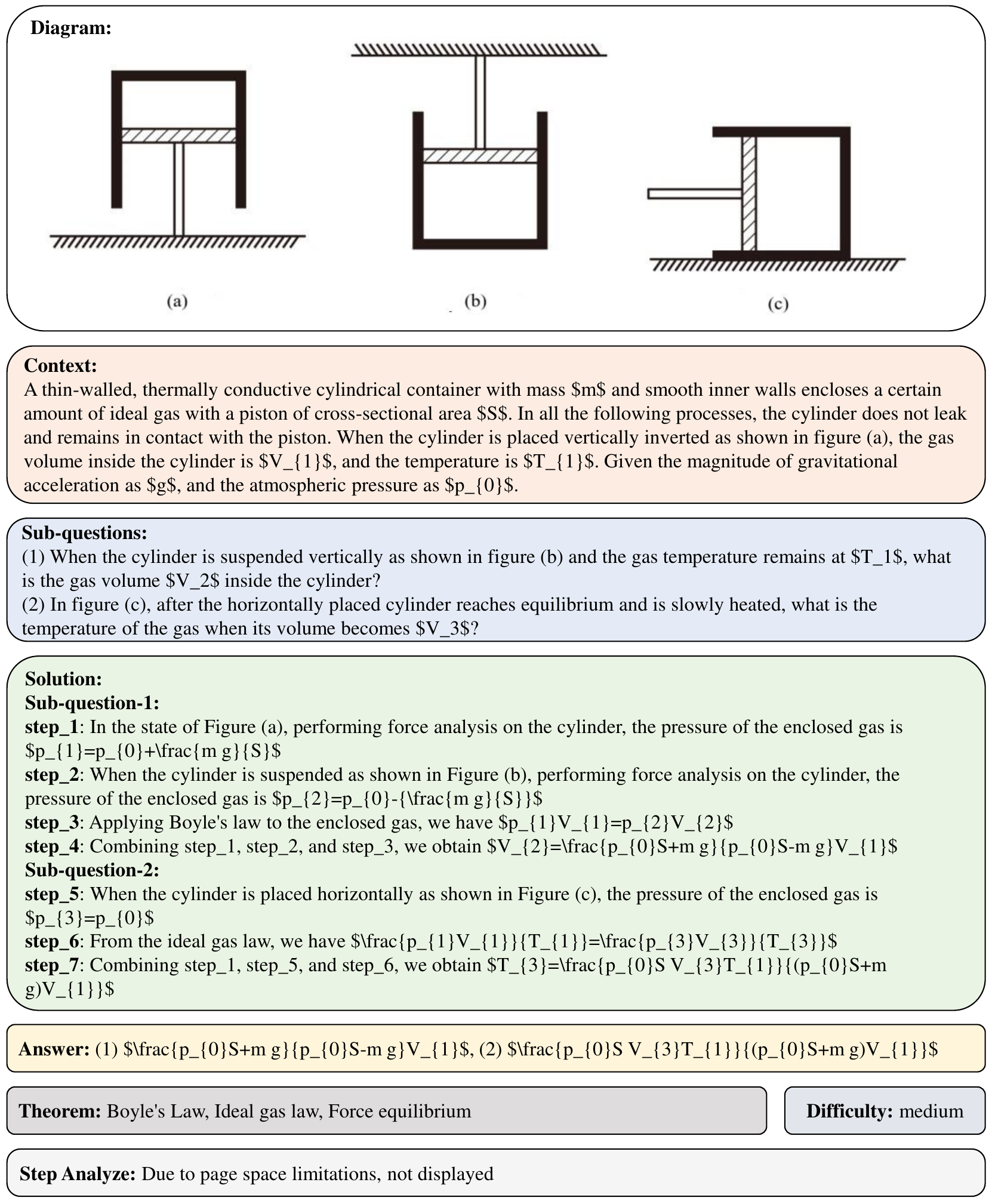}
\caption{A medium example in our benchmark.}
\label{fig:sup_2}
\end{figure*}

\begin{figure*}[t]
\centering
\includegraphics[width=\textwidth]{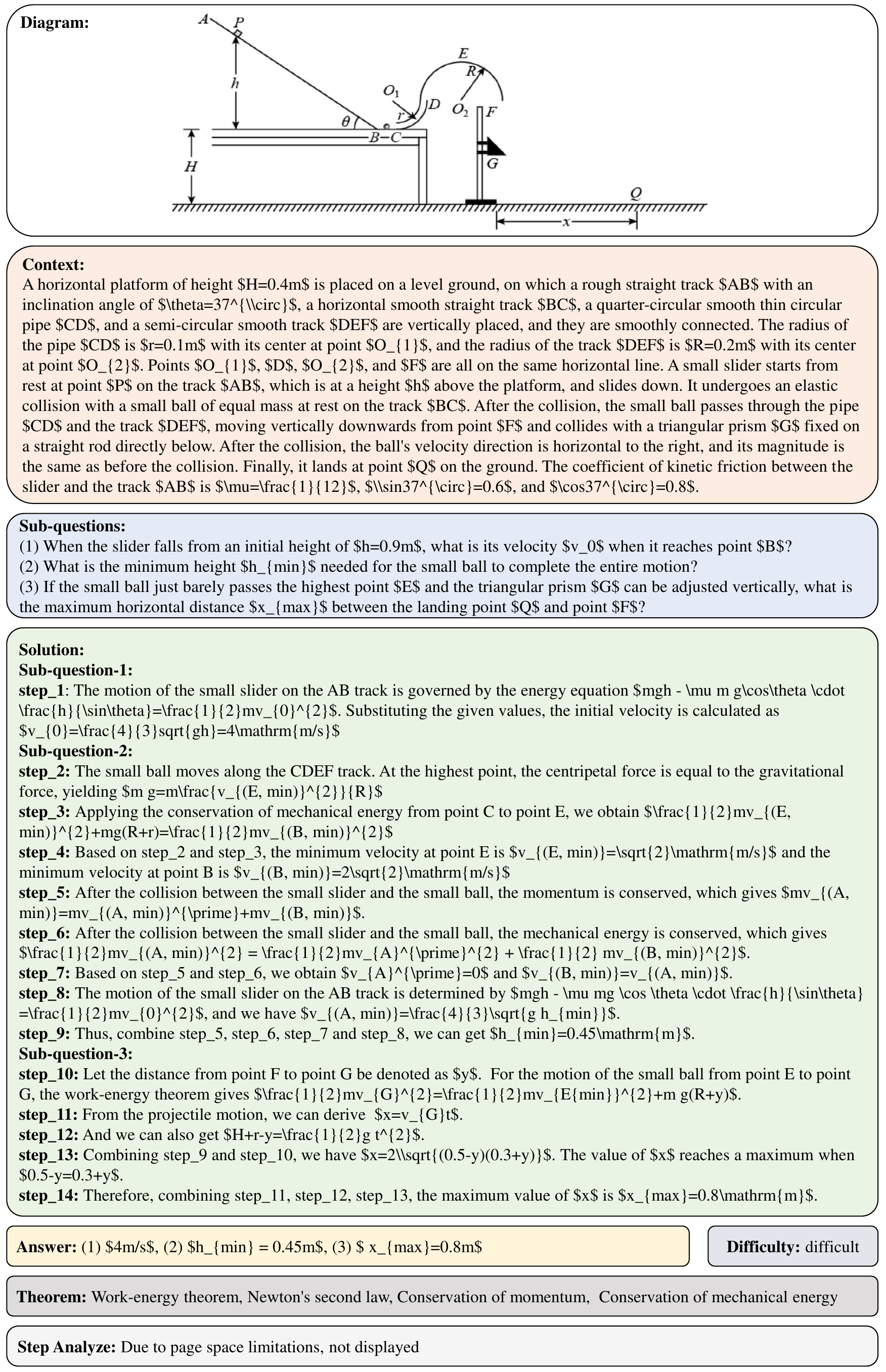}
\vspace{-20pt}
\caption{A hard example in our benchmark.}
\label{fig:sup_3}
\end{figure*}
\section{Example}
\label{Example}
We have provided a representative example for each of the four question difficulty levels—knowledge (Figure \ref{fig:sup_1}), easy (Figure \ref{fig:sup_0}), medium (Figure \ref{fig:sup_2}), and hard (Figure \ref{fig:sup_3}) to serve as a guide.
\par
The knowledge-level problem demonstrates the fundamental application of electromagnetic principles, requiring direct use of basic physics theorems without complex problem-solving steps.
This type of question focuses on testing models' understanding of core concepts and their ability to apply basic formulas.
\par
The easy-level problem involves a straightforward mechanical system with clear physics conditions. 
It requires models to apply basic conservation laws and Newton's laws in a sequential manner, with each step building logically on the previous one. 
The solution path is direct and requires minimal manipulation.
\par
The medium-level problem introduces multiple state changes and requires models to analyze a system under different configurations. 
It combines several physics principles and demands a more sophisticated understanding of how different variables interact.
The solution requires models to track system changes systematically while maintaining consistency in their physics-based reasoning.
\par
The hard-level problem presents a complex mechanical system with multiple connected components and sequential events. 
It requires models to analyze a series of interactions, apply multiple physics principles simultaneously, and consider various constraints throughout the problem-solving process. 
The solution demands both careful physics insight and mathematical rigor, testing models' ability to synthesize different concepts and handle multi-step calculations.
\par
These examples demonstrate the progressive complexity in physics problem-solving across different difficulty levels. From knowledge-level questions testing basic concept application, to hard problems requiring integration of multiple physics principles and sophisticated analysis, each level builds upon the previous one.
This hierarchical structure effectively assesses models' comprehension and problem-solving abilities, ranging from fundamental understanding to advanced physics-based reasoning and mathematical manipulation. The gradual increase in complexity helps evaluate models' mastery of both individual concepts and their ability to synthesize multiple physics principles in complex scenarios.
\begin{table*}[t]
\centering
\caption{Model performance comparisons on the PhysReason-mini benchmark using answer-level evaluation across different input combinations of Questions (Q), Images (I), and Image Captions (IC).
Gemini-2.0-T$^{\dagger}$ and $^{*}$ represent Gemini-2.0-Flash-Thinking-1206 and 0121.}
\begin{tabular}{lcccccc}
\hline
Model & Input & Knowledge & Easy & Medium & Hard & Avg. \\
\hline
\textbf{Non-O-like Models} \\
Qwen2VL-72B & Q, I & 25.40 & 27.00 & 11.4 & 8.5 & 18.07\\
InternVL2.5-78B & Q, I & 37.90 & 20.60 & 18.14 & 7.97 & 21.15\\
GPT-4o & Q, I & 51.12 & 31.95 & 20.75 & 12.54 & 29.09\\
Claude-3.5-Sonnet & Q, I & 49.00 & 40.43 & 23.45 & 12.33 & 31.3\\
Deepseek-V3-671B & Q, IC & 56.60 & 40.97 & 22.22 & 14.61 & 33.6\\
Gemini-2.0-Flash & Q, I  & 67.80 & 52.10 & 40.00 & 23.19 & 46.52\\
Gemini-2.0-Pro & Q, I & 69.32 & 53.67 & 44.98 & 26.24 & 48.55 \\
\hline
\textbf{O-like Models} \\
o1-mini & Q, IC & 54.80 & 30.33 & 15.41 & 7.92 & 27.11 \\
QvQ-72B & Q, I & 51.17 & 37.10 & 29.83 & 22.13 & 35.06 \\
QwQ-32B  & Q, IC & 64.4  & 50.07 & 38.88 & 27.45 & 45.20 \\
Gemini-2.0-T$^{\dagger}$ & Q, I & 71.47 & 49.97 & 36.83 & 22.97 & 45.42 \\
GLM-Zero  & Q, IC & 72.70 & 50.17 & 43.42 & 24.70 & 47.75 \\
o1  & Q, I & 72.47 & 53.37 & 49.31 & 25.32 & 50.12 \\
o3-mini-high  & Q, IC & 71.10 & 63.20 & 47.02 & 31.93 & 53.31\\
Gemini-2.0-T$^{*}$ & Q, I & 76.33 & 56.87 & 51.85 & 32.61 & 54.42 \\
Deepseek-R1 & Q, IC & 85.17 & 60.77 & 47.24 & 33.23 & 56.60 \\
\hline
\end{tabular}
\label{tab:30}
\end{table*}

\section{Evaluation Prompt}
\label{evaluation_prompt}
To systematically evaluate models' mathematical reasoning capabilities, we designed a structured prompt template that follows the zero-shot Chain-of-Thought (CoT) paradigm. 
This template adopts a hierarchical structure comprising image information, problem context, and sequential sub-questions, requiring models to provide standardized step-by-step solutions. The prompt structure consists of the following key components:

\subsection{Input Components}
\begin{itemize}
    \item \textbf{Image Caption:} For models without direct image processing capabilities, we utilize Gemini-2.0-flash to generate image descriptions as supplementary information
    \item \textbf{Context:} Provides the overall background and fundamental information of the problem
    \item \textbf{Sub-questions:} Decomposes complex problems into progressive sub-questions
\end{itemize}

\subsection{Output Specifications}
The template requests a structured output format with the following requirements:
\begin{itemize}
    \item Step-by-step reasoning for each sub-question
    \item Continuous step numbering across sub-questions
    \item One formula and its solution process per step
    \item Mathematical formulas enclosed in LaTeX notation (\$)
\end{itemize}
\par
This design adheres to the zero-shot Chain-of-Thought paradigm, facilitating systematic thinking through explicit step division and standardized output format, which benefits both model reasoning and subsequent performance evaluation. 
The template's flexibility allows it to accommodate pjhysical problems of varying complexity, with adjustable numbers of sub-questions and solution steps based on specific problem requirements.
\section{Details of Experimental Result}
We previously presented only partial model performance benchmarks on PhysReason-mini. 
And we provide a comprehensive performance evaluation across all models, as shown in Table \ref{tab:30}.
\par
\section{Details of Scientific Artifacts}
Our PhysReason benchmark dataset integrates problems from multiple sources: International Physics Olympiad (2008-2021), Chinese National College Entrance Examination (2010-2024), Indian Joint Entrance Examination Advanced (2010-2024), Chinese provincial and municipal mock examination questions (2015-2024), and additional physics problems from Russian sources, totaling over 20,000 unique physics problems from 1,254 PDF documents across 27,874 pages. 
The dataset has been carefully curated to ensure comprehensive coverage while respecting intellectual property rights - all problems are utilized under the CC BY-NC-SA and MIT licenses, and all materials were collected in accordance with fair use principles for educational and research purposes. 
We maintain strict privacy standards with no personally identifiable information included, as all problems are sourced from standardized tests and competition materials. 
The complete benchmark and associated code are made available for research use, requiring users to comply with both the MIT License terms for our implementation and the respective original licenses (CC BY-NC-SA) for the educational materials, thereby ensuring proper attribution and usage rights while promoting academic accessibility.

\section{Details of Computational Experiment}
Our computational experiments were conducted across multiple Large Language Models (LLMs), Vision Language Models (VLMs), and other specialized models. The infrastructure primarily consisted of NVIDIA A800 GPUs for running open-source models. 
For model specifications, we evaluated seventeen models in total, including Qwen2-VL-72B (72 billion parameters), QwQ-32B (32 billion parameters), QvQ-72B (72 billion parameters), InternVL2.5-78B (78 billion parameters), and various other commercial models like GPT-4, Claude-3.5-Sonnet, and Gemini series. 
All experiments were conducted under a zero-shot Chain-of-Thought (CoT) setting to encourage step-by-step reasoning. For the experimental setup, we utilized specific prompts (detailed in supplementary materials) to maintain consistency across all evaluations. The models processed the complete PhysReason benchmark dataset, with the exception of O1 due to API limitations. 
For performance evaluation, we employed both PSAS-A and PSAS-S frameworks, with Deepseek-V3 ultimately selected as the scoring model based on efficiency and performance considerations.
Regarding implementation details, models that couldn't process visual inputs were supplemented with image captions generated by Gemini-2.0-Flash. 
For reproducibility purposes, all prompt templates are provided in the supplementary materials.
Due to the computational cost of the PSAS-S framework, some experiments were conducted using only the PSAS-A framework to maintain efficiency.

\section{Details of human annotators}
For data annotation and evaluation, we engaged four graduate students (including both PhD and Master's students) from engineering disciplines who are also co-authors of this paper. 
All annotators possessed strong backgrounds in both high school and undergraduate physics, making them well-qualified for this task. 
Since the annotators were co-authors actively involved in the research, no formal recruitment process or compensation was required, and they were fully aware of how the data would be used in the study. 
The annotation process focused solely on physics content evaluation and did not involve collecting any personal identifying information or expose annotators to any risks.
As this research involved co-authors analyzing academic content rather than external human subjects, it was determined to be exempt from formal ethics review board approval. 
The annotation work was conducted as part of regular academic research activities within our institution.  
No protected or sensitive demographic information was collected or used in this research.

\section{Details of Ai Assistants In Research Or Writing}
We used Claude-3.5-Sonnet, o1, o3-mini-high, and Deepseek-R1 to help us write code and polish the paper.

\end{document}